\documentclass{article}
\usepackage{iclr2021_conference,times}
\usepackage{amsmath,amsfonts,bm}

\def\eqref#1{equation~\ref{#1}}
\def\1{\bm{1}}

\DeclareMathAlphabet{\mathsfit}{\encodingdefault}{\sfdefault}{m}{sl}
\SetMathAlphabet{\mathsfit}{bold}{\encodingdefault}{\sfdefault}{bx}{n}

\usepackage{hyperref}
\usepackage{url}
\usepackage{graphicx}
\usepackage{amsmath}
\usepackage{amsthm}
\usepackage{booktabs}
\usepackage{subfigure}
\usepackage{multirow}
\usepackage{multicol}
\usepackage[para,online,flushleft]{threeparttable}
\usepackage{lscape}
\usepackage{makecell}
\usepackage{dirtytalk}
\usepackage{amsfonts, bm}
\usepackage{xcolor}
\usepackage{color}
\usepackage{epstopdf}
\usepackage[boxed,ruled]{algorithm2e}
\usepackage{setspace}
\usepackage{enumitem}
\usepackage{float}
\title{High-Robustness, Low-Transferability Fingerprinting of Neural Networks}

\author{Siyue Wang$^{1}$, Xiao Wang$^{2}$, Pin-Yu Chen$^{3}$, Pu Zhao$^{1}$ \& Xue Lin$^{1}$ \\

1. Northeastern University \ 2. Boston University  \ 3. IBM Research\\
\texttt{\{wang.siy,zhao.pu,xue.lin\}@northeastern.edu \ kxw@bu.edu \ pin-yu.chen@ibm.com} \\
}

\iclrfinalcopy % Uncomment for camera-ready version, but NOT for submission.
\begin{document}

\maketitle

\begin{abstract}
This paper proposes  \textit{Characteristic Examples} for effectively fingerprinting deep neural networks, featuring high-robustness to the base model against model pruning as well as low-transferability to unassociated models. 
This is the first work taking both robustness and transferability into consideration for generating realistic fingerprints, whereas current methods lack practical assumptions and may incur large false positive rates. 
To achieve better trade-off between robustness and transferability, we propose three kinds of characteristic examples: \textit{vanilla C-examples},  \textit{RC-examples}, and \textit{LTRC-example}, to derive fingerprints from the original base model.
To fairly characterize the trade-off between robustness and transferability, we propose \textit{Uniqueness Score}, a comprehensive metric that measures the difference between robustness and transferability, which also serves as an indicator to the false alarm problem. 
% Extensive experiments demonstrate that the proposed characteristic examples can achieve superior performance when compared with existing fingerprinting methods. In particular, for VGG ImageNet  models, using LTRC-examples gives $4\times$ higher uniqueness score than the baseline method and  does not incur any false positives.
\end{abstract}

\vspace{-2mm}
\section{Introduction}
\vspace{-1mm}

%With the rapid development of machine learning and artificial intelligence, 
Tremendous efforts have been spent on developing state-of-the-art machine learning models e.g., deep neural networks (DNNs).
%can be , and therefore it is of utmost importance to be able to claim the ownership of a well-trained model and its derived versions (e.g. pruned models). 
For instance, the cost of training current state-of-the-art transformer based language model, GPT-3 \cite{brown2020language}, is estimated to be at least 4.6 million US dollars\footnote{\url{https://bdtechtalks.com/2020/08/17/openai-gpt-3-commercial-ai}}. 
Imagine an unethical model thief purposely pruned the pre-trained GPT-3 model and attempted to claim the ownership of the resulting compressed model. 
We need to answer the question of ``how to protect intellectual property for DNN models and reliably identify model ownership?''
%is literally worth million dollars.

Another motivating example is the surging trend of broad usage of neural network models across applications in cloud-based or embedded systems.
For model owners deploying a model on the cloud, it is essential for them to verify the identity of the model to make sure that the model has not been tampered or replaced. 
% Towards this direction, extensive research have been made to protect the IP of the neural network from different perspectives, which can be regard as fingerprinting/watermarking using weights embedding and image examples. 
However, most of these current methods for DNN IP protection require intervention in training phase, which may cause performance degradation of the DNN (i.e., accuracy drop) and leave hidden danger of adversary to attack the DNN (i.e., backdoor attacks). Meanwhile, existing works often overlook the false positive problem of the DNN (i.e., mistakenly claiming the ownership of irrelevant models), which is of practical importance when designing fingerprints. 

To better address the aforementioned limitations, this work proposes a novel approach to fingerprinting DNNs using \textit{Characteristic Examples} (C-examples). Its advantages lies in that (i) its generation process does not intervene with the training phase; and (ii) it does not require any realistic data from the training/testing set. By applying uniform random noise to the weights of the neural network with the combination of gradient mean descending technique, the proposed C-examples achieve high-robustness to the resulting models pruned from the base model where the fingerprints are extracted. When further equipped with a high-pass filter in the frequency domain of image data during the generation process, C-examples attain low-transferability to other models that are different from the base model. 
Extensive experiments demonstrate that the proposed characteristic examples can achieve superior performance when compared with existing fingerprinting methods. In particular, for VGG ImageNet  models, using LTRC-examples gives $4\times$ higher uniqueness score than the baseline method and  does not incur any false positives.

% Below we summarize our main contributions.
% \begin{itemize}[leftmargin=*]
% \item We propose a novel and practical fingerprinting method called \textit{C-examples} that achieves better robustness and transferability trade-off than current DNN fingerprinting methods without intervening with the training phase nor the dataset. In particular, we develop three kinds of C-examples: \textit{vanilla C-examples}, \textit{RC-examples}, and \textit{LTRC-examples} to further improve fingerprinting performance.

% \item To better evaluate the trade-off between robustness and transferability, we propose a novel metric called \textit{Uniqueness Score} that quantifies the utility of fingerprinting. The uniqueness scores of the proposed C-examples outperform other fingerprinting methods by a large margin. Specifically, LTRC-examples gives $4\times$ higher uniqueness score than the baseline. 

% \item This is the first work that thoroughly considers the false alarm problem in designing fingerprints. Our mechanisms featuring high-robustness and low-transferability can significantly decrease the false positive rate and achieve nearly perfect AUC and F1-score with LTRC-examples.
% \end{itemize}
\begin{figure*}[htbp!]
    \centering
    \vspace{-2mm}
    \includegraphics[width=0.85\textwidth]{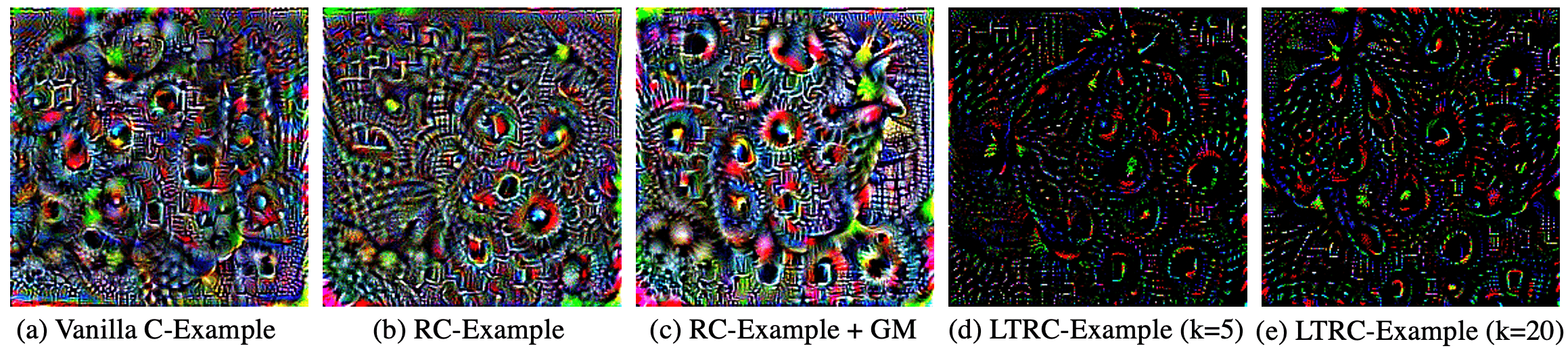}
    \vspace{-2mm}
    \caption{\textbf{Characteristic Examples Visualized using Different Generation Process.}  The label assigned to all these image is “strawberry”.
    }
    \label{fig:visualize_img1}
\end{figure*}

\vspace{-2mm}
\section{Characteristic Examples for Neural Network Fingerprinting}
\vspace{-1mm}
Our fingerprinting methods are introduced in the context of image classification task by DNNs.
We consider three types of DNN models that are of interest in %during generating
C-examples.
\textcircled{\small{1}} \textit{Base Model}: the pre-trained model to fulfill some designated task, such as image classification. 
\textcircled{\small{2}} \textit{Pruned Models}: the models pruned from the base model and implemented on the edge devices for inference execution.  
\textcircled{\small{3}} \textit{Other Models}: any other models that are neither \textcircled{\small{1}} nor \textcircled{\small{2}}. 
Our goal is to design C-examples that are both robust to \textcircled{\small{2}} Pruned Models and exhibiting low-transferability on \textcircled{\small{3}} Other Models.

\subsection{Proposed C-Examples}\label{sec:C-examples}
% In this section, our main methods are introduced in the context of image classification task by DNNs.
% We stress, however, that the proposed approach can be generalized to other types of tasks, data, and classification models.
%In this section, we propose our framework of generating characteristic examples as a DNN fingerprinting method. We mainly focus on the image classification tasks.
Let $\mathbf{x}\in\mathbb{R}^{3\times H \times W}$ denote a colored RGB image,  
%where $H$ and $W$ are the image height and width, respectively.
where the pixel values of $\mathbf{x}$ are scaled to $[0,1]$ for  mathematical simplicity.
$F_{\theta}$  denotes the pre-trained Base Model, which outputs $\bm{y}={F_{\theta}}(\mathbf{x})$ as a probability distribution for a total of $M$ classes.
The element ${{y}_i}$ represents the probability that an input $\mathbf{x}$ belongs to the $i$-th class. 
% The Base Model $F_{\theta}$ parameterized with $\theta$ is pre-trained.
% Then, C-examples are generated from $F_{\theta}$.
If we are given with a subset $\{l_1,l_2,\dots,l_P\}$ of $P$ labels randomly chosen from the labels of  training dataset, then a set of $\eta$-optimal ($\eta$ set to $1\times 10^{-6}$ to guarantee the convergence) RC-examples ${X^*}$ can be characterized as:
{\small
\begin{equation}\label{eqn:RCE_formulation}
{X^*} = \Big\{(\mathbf{x},l)\big|
\text{Loss}_{\theta}(\mathbf{x},l) < \eta, \mathbf{x}\in [0,1]^n \Big\}\\.
\end{equation}}%
% We set $\eta$ to   $1\times 10^{-6}$ in order to guarantee the convergence of the generation.

The $\text{Loss}_{\theta}(\cdot)$ denotes the loss function of $F_{\theta}$.
A C-example $\mathbf{x}$ minimizing the loss for a specified label $l$  should satisfy the  above constraint. We use a random seed to generate a C-example, 
% and therefore the generated C-examples are distinct from natural images for the human perception.
where a vanilla version C-example is shown in Figure \ref{fig:visualize_img1} (a).

We choose to use the projected gradient decent (PGD) algorithm  \cite{lin2007projected,kurakin2017adversarial,kurakin2016adversarial,madry2018towards}, which has been widely used as a general approach for  solving  constrained optimization problems.
Then the C-example generation problem (\ref{eqn:RCE_formulation}) can be solved 
%with the PGD algorithm  
as:
{\small\begin{equation}\label{eqn:pgd}
\mathbf{x}^{t+1} = {\rm \textbf{Clip}}
\left(\mathbf{x}^t-\alpha\cdot\text{sign}(\nabla_{\mathbf{x}}\text{Loss}_{\theta}(\mathbf{x}^t,l))\right),
\end{equation}}%
where $t$ is the iteration step index; $\mathbf{x}^0$ is the random starting point;
$\alpha$ is the step size; $\text{sign}(\cdot)$ returns the element-wise sign of a vector;  $\nabla_\mathbf{x}(\cdot)$ calculates gradients; and $\textbf{Clip}(\cdot)$ denotes the clipping operation to satisfy the $\mathbf{x}\in [0,1]^n$ constraint.
In summary, the PGD algorithm generates a C-example by iteratively making updates based on the gradients and then clipping into the $\ell_\infty$-ball.
% ($[0,1]^n$).

\subsection{C-Examples with Enhanced Robustness}\label{sec:RC-examples}

% All the existing works \cite{le2019adversarial,he2019sensitive} 
% perform fingerprinting or watermarking for a neural network as it is.
% They can not differentiate the (benign) model compression -- an essential step in implementing neural network models for on-the-edge inference execution, from other adversarial model perturbations.
One of the major limitations of existing works \cite{le2019adversarial,he2019sensitive} (detailed in Appendix \ref{background}) for fingerprinting or watermarking DNNs is that they can not differentiate the (benign) model compression.
% -- an essential step in implementing neural network models for on-the-edge inference execution, from other adversarial model perturbations.
Here, we tackle this challenge by improving the robustness of C-examples on \textcircled{\small{2}} Pruned Models by \emph{\textbf{Robust C-examples (RC-examples)}},
% It means that the C-examples generated from the  \textcircled{\small{1}} Base Model should also preserve a high test accuracy on \textcircled{\small{2}} Pruned Models that are derived from the \textcircled{\small{1}} Base Model. 
% To achieve this, we propose an enhancement named \emph{\textbf{Robust C-examples (RC-examples)}} over the vanilla version proposed in Section \ref{sec:C-examples},
by adding noise  bounded by $\delta$ to the neural network parameter $\theta$ to mimic the model perturbation due to the model compression. 
Here the loss is changed to $\text{Loss}_{\theta+\Delta}$, where $\Delta$ presents the uniformly distributed weight perturbations within $[-\delta,\delta]$.
% We set $\delta$ as 0.001, 0.003, 0.005, 0.007 for ImageNet  and 0.01, 0.03, 0.05, 0.07 for CIFAR-10, respectively, in the experiments.

\emph{\textbf{Further Robustness Enhancement with Gradient Mean (GM).}}
Furthermore, motivated by the EOT method \cite{athalye2018obfuscated,wang2019protecting} towards stronger adversarial attacks, the proposed RC-examples can be further enhanced by calculating the mean of the input gradients in each iteration step.
When computing input gradient, we sample input gradients for $q=10$ times and use the mean of gradients in  each iteration  step of generating RC-examples.

\begin{figure}[t]
    \centering
    \vspace{-3mm}
    \includegraphics[width=0.95\textwidth]{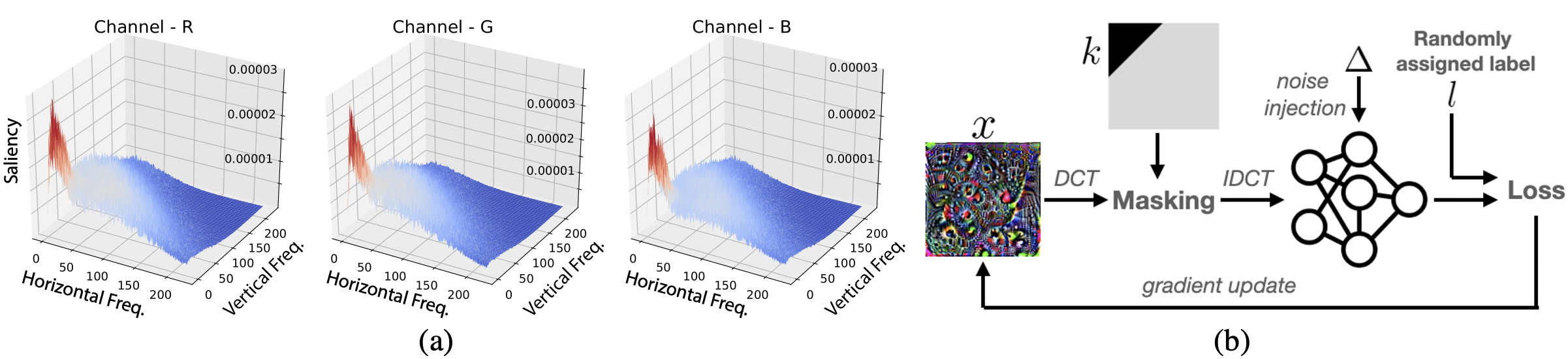}
    \vspace{-3mm}
    \caption{\textbf{(a) Saliency map} of three color channels averaged over 1000 images from  ImageNet demonstrating the  absolute gradients of the base model classification loss  w.r.t. the frequencies obtained from the DCT of input images. \textbf{(b) A system diagram } of generating LTRC-example.
    } 
    \vspace{-3mm}
    \label{fig:fre_show}
\end{figure}

\subsection{RC-Examples with Low-Transferability}

% \begin{figure}[t]
%     \centering
%     \includegraphics[width=0.4\textwidth]{Fig/system4.png}
%     \caption{\textbf{A System Diagram of Generating LTRC-example.}
%     }
%     \label{fig:system_show}
% \end{figure}

% Besides enhancing the robustness of C-examples on \textcircled{\small{2}} Pruned Models, it is desirable to exhibit low-transferability to \textcircled{\small{3}} Other Models. In other words, the C-examples should be able to distinguish the implemented inference model on the edge device, if it is an \textcircled{\small{3}} Other Model.
We further improve the RC-examples proposed in Section \ref{sec:RC-examples} by enforcing low-transferability to \textcircled{\small{3}} Other Models, i.e., we propose the \emph{\textbf{Low-Transferability RC-examples (LTRC-examples)}}.
In this way, we can improve the capability of C-examples in detection for \emph{false positive} cases, where positive means claiming the model ownership as ours in IP protection. 
% The frequency analysis \cite{guo2018low,sharma2019effectiveness,cheng2019improving} suggests that low frequency components can improve transferability of adversarial examples. 
% Inspired by that, we propose to leverage high frequency components to achieve C-examples with low-transferability.
Specifically, we apply a frequency mask on the {\it{Discrete Cosine Transform}} (DCT) \cite{rao2014discrete} to implement a high-pass filter in the frequency domain of the C-example, detailed in Appendix \ref{frequency_analysis} .

For most images, we found that the low-frequency components are mostly salient for deep learning classifiers. 
As shown in Figure~\ref{fig:fre_show}(a),  the low-frequency components  in the red area around (0,0) have a larger contribution (with larger gradients) to the classification loss.
Inspired by this phenomenon,
% that  the low-frequencies play a more important role in   classifications  and therefore are more transferable, 
we believe that filtering out these components can effectively lower the fingerprints transferability.
To demonstrate this, we design the high-pass frequency mask as shown in Figure \ref{fig:fre_show}(b), where the high-frequency band size $k$  controls the  range of the filtered low-frequency components. 
% The frequency mask is designed to be a 2D matrix with elements being either 0 or 1, i.e., $\mathbf{m}\in \{0,1\}^{H \times W}$ which performs element-wise product with the DCT of C-example.
% At each iteration step to generate the fingerprints, the high-pass mask sets the low-frequency components to 0, i.e., $ \omega_{(i,j)}=0$  if $ 1 \leq i+j \leq k$, while keeping the rest of the high-frequency components.
% On ImageNet dataset with  mask size $H = W = 224$ ($H = W = 32$ for CIFAR-10 dataset), the high-frequency band size $k=20$ leads to $\frac{1}{2}\times20^2 / 224^2 \approx 0.4\%$ of the frequency components set to 0.

By using the high-pass frequency mask, the LTRC-example at the $(t+1)$-th iteration step can be derived by:
% \begin{equation}\label{eqn:freq_mask}
% \mathbf{x}^{t+1} = \rm {FreqMask}\Big\{
% {\rm \textbf{Clip}}
% \left(\mathbf{x}^t-\alpha\cdot\text{sign}(\nabla_{\mathbf{x}}\text{Loss}_{\theta+\Delta} (\mathbf{x}^t,l))\right)\Big\},
% \end{equation}
{\small
\begin{equation}\label{eqn:freq_mask}
\mathbf{x}^{t+1} = \rm {HighPass}\Big\{
{\rm \textbf{Clip}}
\left(\mathbf{x}^t-\alpha\cdot\text{sign}(\nabla_{\mathbf{x}}\text{Loss}_{\theta+\Delta} (\mathbf{x}^t,l))\right)\Big\},
\end{equation}}%
where the HighPass filter is defined as:
{\small\begin{equation}
\rm {HighPass}(\cdot) = IDCT({FrequencyMask} (DCT(\cdot))).
\end{equation}}%

\vspace{-3mm}
\section{Performance Evaluation}
\vspace{-2mm}
% In our experiment, we use the accuracy of the C-examples on the pruned model to indicate its robustness and the accuracy on the  variant model (with  similar functionality to the base model, e.g. VGG-19 model to the base VGG-16 model) to indicate its transferabilty. Originally, the accuracy of all kinds of C-examples on the base model is  100\% during generation.
% To effectively evaluate the trade-off between robustness of the pruned models and transferability to other variant models,  we define the difference between the robustness and transferability %to the worst-case model 
% as \textit{Uniqueness Score} ($\textit{Uniqueness Score } = \textit{Robustness} - \textit{Transferability} $),
% % \PY{(Somehow I feel uniqueness = robustness - tranferability is easier to be understood)}
% where higher \textit{Uniqueness Score} means the C-examples are more robust to pruned models and less transferable to variant models.
% %achieve better trade-off between robustness and transferability, and vice versa.
% Intuitively, a better fingerprint method should achieve higher uniqueness score.
% \textit{Uniqueness Score} can also be used to indicate the false positive problem, i.e., if \textit{Uniqueness Score} is negative, the corresponding fingerprint method is prone to make false model claims.

The implementation details and comparative methods for our experiments are summarized in Appendix \ref{implimentation_details} and \ref{comparative_methods}. 
In our experiment, we use the accuracy of the C-examples on the \textcircled{\small{2}} Pruned Model to indicate its robustness and the accuracy on the \textcircled{\small{3}} Other Model to indicate its transferabilty. 
% Originally, the accuracy of all kinds of C-examples on the base model is  100\% during generation.
To evaluate the trade-off between robustness  of the pruned models and transferability  to other variant models,  we define the difference between the robustness and transferability %to the worst-case model 
as \textit{Uniqueness Score} ($\textit{Uniqueness Score } = \textit{Robustness} - \textit{Transferability} $),
where higher \textit{Uniqueness Score} represents more robustness to pruned models and less transferability to variant models.
%achieve better trade-off between robustness and transferability, and vice versa. 
% Intuitively, a better fingerprint method should achieve higher uniqueness score.
% \textit{Uniqueness Score} can also be used to indicate the false positive problem, i.e., if \textit{Uniqueness Score} is negative, the corresponding fingerprint method is prone to make false model claims.
We demonstrate the effectiveness of proposed methods on different pruned models for evaluating robustness and VGG-19 (worst-case transferability to VGG-16) for testing transferability.
% For testing transferability to other variant models, as VGG-19  is the most similar architecture to VGG-16 and more transferable for fingerprints  generated on VGG-16, 
% we only report the  transferability on VGG-19 and omit the transferability results on other models such as ResNet \say{Family} or DenseNet \say{Family}. Note that the transferability to other models should be lower than  VGG-19, leading to better performance with higher uniqueness score. 

\begin{figure*}[htbp]
    \centering
    \vspace{-3mm}
    \includegraphics[width=0.92\textwidth]{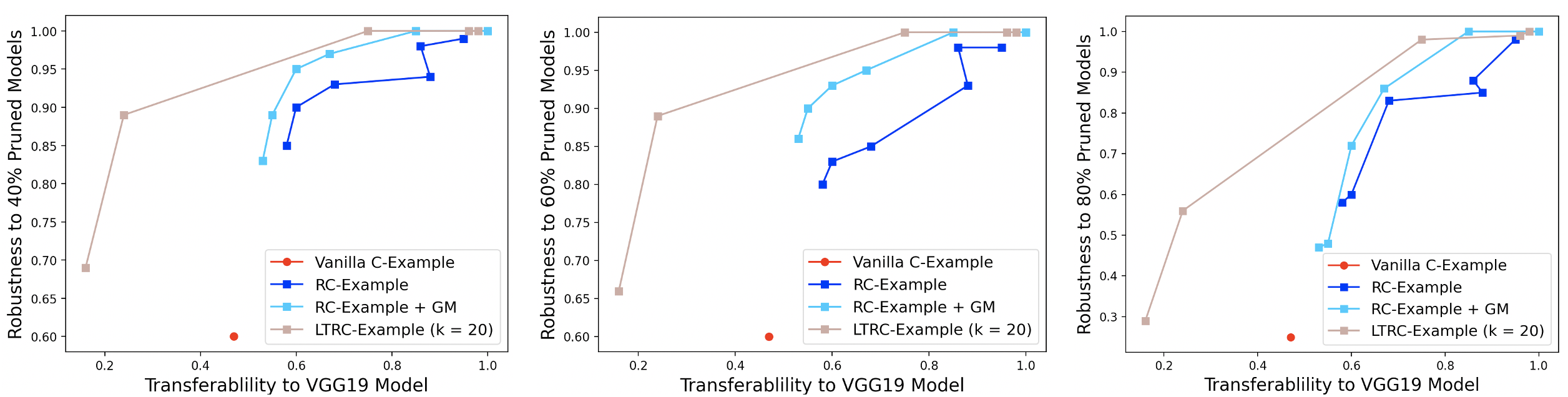}
    \vspace{-2mm}
    \caption{\textbf{Visualization of the Trade-off Curve between Transferability and Robustness.} Base Model is Pruned with 40\%, 60\%, and 80\% Pruning Ratios. 
    }
    \label{fig:visualize_curve}
\end{figure*}
\vspace{-2mm}

% \subsection{Uniqueness Evaluation}\label{uniqueness_eval}

We plot and visualize the trade-off curve between robustness to the pruned models and transferability to variant models in Figure \ref{fig:visualize_curve}, and Table \ref{table_general} further summarizes the corresponding \textit{Uniqueness Score} with respect to each method. For better comparison, we only take our best choice of $k = 20$.
% for LTRC-examples. 
% More details can be found in Table \ref{main:table_ablation_0.001}  for the ablation study of  $k$ value.
We summarize our findings from experiments as follows:
\begin{enumerate}[label=\arabic*,leftmargin=*, noitemsep]

\item For evaluating the trade-off between robustness and transferability, as shown in Figure \ref{fig:visualize_curve}, 
both RC-examples, RC-examples+GM, and LTRC-examples clearly outperforms the baseline vanilla C-examples as fingerprinting methods. By comparing RC-examples and RC-examples+GM, applying GM to the input gradients can significantly help with the fingerprinting performance on both robustness and transferability. 
The proposed LTRC-examples clearly outperforms C-examples, RC-examples, and RC-examples+GM for all pruned models, as LTRC-example applies both random perturbations to the weights and high-pass filters to remove the high-transferable low-frequency components during generation.
% Specifically,  as the classification mainly relies on  low-frequency components, LTRC-example can significantly decrease its transferbility to other models by applying the high-pass frequency mask. 

\item %For evaluating the uniqueness score of different kinds of C-examples, 
As shown in Table \ref{table_general}, uniqueness scores of RC-examples, RC-examples+GM, and LTRC-examples are higher than that of the baseline vanilla C-examples. We notice that C-examples  suffer from negative uniqueness scores due to their high transferability to other models  when the pruning ratios are 70\% and 80\%. 
We can observe that LTRC-examples with $\delta=0.001$ achieve the best uniqueness scores with relatively large margins (about 1.9X, 2.1X, and 5X that of the RC-examples+GM, RC-examples, and C-examples).

\item In general, for a given method with fixed $\delta$, the uniqueness score decreases if the pruning ratio increases since  larger pruning ratio degrades the test accuracy, leading to weaker model functionalities with less robustness after pruning. 
%. The reason is that in this case the transferability is fixed and models pruned with larger pruning ratio keep fewer functionalities that is the same with the pre-trained model, i.e. test accuracy.
Meanwhile, we  observe that with increasing $\delta$, there are more uncertainty in the model with larger random perturbations, leading to more general C-examples to incorporate larger uncertainty.
%which increases the robustness of fingerprints to pruned models, 
Thus they become more transferable to other variant models, resulting in increasing  transferability and decreasing uniqueness score.
%the transferability of the fingerprints to variant model will also be increased, causing negative effect to uniqueness score in general. 
For example, for LTRC-examples with $\delta=0.001$, with 40\% pruned model, the uniqueness score is 65 while it becomes 2 with $\delta=0.007$.
\end{enumerate}

\begin{table*}[htbp!]\small
 \centering
 \vspace{-3mm}
  \caption{\textbf{Uniqueness Score of C-examples on Implemented Models by Different Weight Pruning on the Base VGG-16 model with ImageNet Dataset:} The base model has 70.85\% top 1 accuracy and 90.10\% top 5 accuracy. The base model is pruned by unstructured pruning \cite{han2015learning} with various pruning ratio, where it is pruned for 5 times at each pruning ratio with average accuracy degradation for pruning ratio 40\% to 80\% are 0.26\%, 0.45\%, 0.38\%, 0.61\%, and 0.97\%, respectively. We choose one representative setting for LTRC-examples with $k=20$.  The robustness at each pruning ratio can be obtained by the summation of \textit{Uniqueness Score} and transferability.}
  %is reported by the addition of \textit{Uniqueness Score} and accuracy of the worst case VGG-19 model representing the transferability of each group of C-examples. } 
  \label{table_general}
  \scalebox{0.75}[0.75]{
  \begin{threeparttable}
    \begin{tabular}{ccccccccc}
    \hline
    \multirow{2}{*}{Method} & \multirow{2}{*}{$\delta$} & Base Model & Transferability & \multicolumn{5}{c}{Uniqueness Score (\%)} \\
    &  & \textbf{VGG-16} (\%) & to \textbf{VGG-19} (\%) & 40\% Pruned  & 50\% Pruned & 60\% Pruned & 70\% Pruned  & 80\% Pruned  \\ \hline 
     Vanilla C-Example &  0 & 100 & 47 & +13 & +13 & +13 & -5 & -22 \\ \cline{1-9} 
     \multirow{4}{*}{RC-Example} &  0.001 &100  & 60 & +30 & +30 & +23 & +22   & +0 \\ 
     &  0.003 & 100 & 88 & +6  & +11 & +5 & +2 & -3 \\ 
     &  0.005 & 100 & 86 & +12 & +12 & +12 & +9 & +2 \\ 
     &  0.007 & 100 & 95 & +4  & +4 & +3 & +2 & +3 \\ \cline{1-9} 
     \multirow{4}{*}{RC-Example+GM}  
     &  0.001 & 100 & 55 & +34 & +37   &  +35 & +23 & -7 \\
     &  0.003 & 100 & 67 & +30  &  +38 & +38  & +21  &  +19 \\ 
     &  0.005 & 100 & 85 & +15   & +15   & +15   & +15  &  +15  \\ 
     &  0.007 & 100 & 100 & +0 & +0  & +0 & +0 & +0  \\ \cline{1-9} 
    %  \multirow{4}{*}{LT-RCE(k = 5)）} &  & 0.001 & 100\% & 26\% & 79\% & 78\%   &  88\% & 71\% & 40\%\\ 
    %  &  & 0.003 & 100\% & 68\% & 94\%  &  96\% & 96\%  & 94\%  &  80\%  \\ 
    %  &  & 0.005 & 100\% &90\% & 98\%   & 98\%  & 96\%  & 95\%  & 86\%  \\ 
    %  &  & 0.007 & 100\% & 93\% & 100\% &100\%  & 100\% & 100\% & 99\%  \\ \cline{1-10}
    %  & \multirow{4}{*}{LT-RCE(k = 10)）} & 0.001 &  &  &  &  &  &  &  \\ 
    %  &  & 0.003 &  &  &  &  &  &  &  \\ 
    %  &  & 0.005 &  &  &  &  &  &  &  \\ 
    %  &  & 0.007 &  &  &  &  &  &  &  \\ \cline{2-10}
    %  & \multirow{4}{*}{LT-RCE(k = 20)）} & 0.001 &  &  &  &  &  &  &  \\ 
    %  &  & 0.003 &  &  &  &  &  &  &  \\ 
    %  &  & 0.005 &  &  &  &  &  &  &  \\ 
    %  &  & 0.007 &  &  &  &  &  &  &  \\ \cline{2-10}
     \multirow{5}{*}{LTRC-Example} 
     &  0 & 100 & 16 & +53  & +51 & +50 & +23  & +13  \\ 
     & 0.001 & 100 & 24 & \textbf{+65} & \textbf{+65} & \textbf{+65} & \textbf{+58} & \textbf{+32} \\ 
     &  0.003 & 100 & 75 & +25  & +25 & +25 & +25 & +23  \\  
     &  0.005 & 100 & 96 & +4 & +4 & +4 & +4 & +3\\  
     &  0.007 &100  & 98 & +2 & +2  & +2 & +2 & +2 \\ \hline
    \end{tabular}
  \begin{tablenotes}
          The experiment is evaluated on 100 C-examples generated from VGG-16.
  \end{tablenotes}
\end{threeparttable}}
\vspace{-2mm}
\end{table*}
\vspace{-1mm}
\section{Conclusion}
\vspace{-2mm}
% We proposed \textit{Characteristic Examples} and \textit{Uniqueness Score} for effective fingerprinting of DNNs that take both robustness and transferability into consideration. Extensive experiments demonstrate that the proposed methods have superior performance in achieving high-robustness and low-transferability than current watermarking/fingerprinting methods. 

Towards achieving high-robustness and low-transferability for fingerprinting DNNs,
we design three kinds of characteristic examples with increasing performance by applying random noise to the model parameters and using a high-pass filter to remove low-frequency components. To fairly characterize the trade-off between robustness and transferability, we propose an evaluation metric named \textit{Uniqueness Score}.  Extensive experiments demonstrate that the proposed methods have superior performance in achieving high-robustness and low-transferability than current watermarking/fingerprinting methods. 

\section*{Acknowledgements}
This   research   is   partially   funded   by   National   Science Foundation CNS-1929300.

\bibliography{iclr2021_conference}
\bibliographystyle{iclr2021_conference}

\newpage
\appendix
\section{Appendix}
\subsection{Background and Related Work}
\label{background}
% \subsection{IP Protection of Deep Neural Networks}

DNN watermarking / fingerprinting methods for the DNN intellectual property protection and model integrity verification can be classified as two main categories:  (1) DNN watermarking or fingerprinting by weights embedding;
% \cite{uchida2017embedding,rouhani2019deepsigns,chen2019deepmarks,fan2019rethinking}
 (2) Watermarking or fingerprinting using image samples.
% \cite{adi2018turning,guo2018watermarking,namba2019robust,he2019sensitive,le2019adversarial,lukas2019deep}.

DNN watermarking following the first approach embeds watermarks into the model weight parameters %through training from scratch, retraining, distillation, and 
which requires white-box access to the model to be tested. Towards this approach, Uchida et al. takes the first step to investigate the DNN watermarking by embedding a watermark in model weight parameters, using a parameter regularizer \cite{uchida2017embedding}. 
Later on Rouhani et al. propose an IP protection framework that enables the insertion of digital watermarks in the target DNN model before distributing the model \cite{rouhani2019deepsigns}. Other works proposed by Chen et al. \cite{chen2019deepmarks} and Fan et al. \cite{fan2019rethinking} also contribute towards this approach.

The second approach extracts the watermarks by using a set of image samples. This line of work includes watermarking by using DNN backdoor attacks \cite{gu2017badnets} to embed watermarks into the DNN model representation while using trigger images to test intellectual property infringement \cite{adi2018turning,guo2018watermarking,namba2019robust}. Another direction is to extracts adversarial examples \cite{le2019adversarial,lukas2019deep} or sensitive examples \cite{he2019sensitive} from a DNN as its fingerprints. 

% \subsection{Frequency Components and Transferability}
% It has been exploited in the area of signal processing and image compression that  most of the critical content-defining information in natural images lies in the low end of the frequency spectrum \cite{wallace1992jpeg}.
% Based on this exploration, the relationship between frequency components of the images and its transferability have been discussed recently to generate adversarial examples as an attack to the neural networks with low frequency adversarial perturbations in order to achieve high-transferability \cite{guo2018low,sharma2019effectiveness}.
% Specifically, by utilizing the well-known discrete cosine transform (DCT), Sharma et al. propose a systematic experiments to evaluate the effectiveness of the low frequency adversarial perturbations by manipulating specific frequency components. They show that the application of using a low-pass filter in the frequency
% domain of the perturbation can effectively improve the transferability thus lead to higher attack success rate, the same phenomenon is also discussed in \cite{guo2018low}.

\subsection{Frequency Analysis}
\label{frequency_analysis}
The frequency analysis \cite{guo2018low,sharma2019effectiveness,cheng2019improving} suggests that low frequency components can improve transferability of adversarial examples. 
Inspired by that, we propose to leverage high frequency components to achieve C-examples with low-transferability.
Specifically, we apply a frequency mask on the {\it{Discrete Cosine Transform}} (DCT) \cite{rao2014discrete} to implement a high-pass filter in the frequency domain of the C-example.

As an important tool in signal processing, the DCT decomposes a given  signal into cosine functions oscillating at different frequencies and amplitudes. 
For a 2D image, the DCT performed as $\rm \omega = DCT (\mathbf{x})$ can transform the image $\mathbf{x}$ into the frequency domain,  and $\omega_{(i,j)}$ is the magnitude of its corresponding cosine functions with the values of $i$ and $j$ representing frequencies, where smaller values mean lower frequencies. 
The DCT is  invertible, and the Inverse DCT (IDCT) is denoted as $\rm \mathbf{x} = IDCT (\omega)$. 
Note that here we apply DCT and IDCT for different color channels independently.

For most ImageNet images, we found that the low-frequency components are mostly salient for deep learning classifiers. 
As shown in Figure~\ref{fig:fre_show},  the low-frequency components  in the red area around (0,0) have a larger contribution (with larger gradients) to the classification loss.
Inspired by this phenomenon that  the low-frequencies play a more important role in   classifications  and therefore are more transferable, we believe that filtering out these components can effectively lower the fingerprints transferability.
To demonstrate this, we design the high-pass frequency mask as shown in Figure \ref{fig:fre_show}(b), where the high-frequency band size $k$  controls the  range of the filtered low-frequency components. 
The frequency mask is designed to be a 2D matrix with elements being either 0 or 1, i.e., $\mathbf{m}\in \{0,1\}^{H \times W}$ which performs element-wise product with the DCT of C-example.
At each iteration step to generate the fingerprints, the high-pass mask sets the low-frequency components to 0, i.e., $ \omega_{(i,j)}=0$  if $ 1 \leq i+j \leq k$, while keeping the rest of the high-frequency components.
On ImageNet dataset with  mask size $H = W = 224$ ($H = W = 32$ for CIFAR-10 dataset), the high-frequency band size $k=20$ leads to $\frac{1}{2}\times20^2 / 224^2 \approx 0.4\%$ of the frequency components set to 0.

\subsection{Implementation Details}
\label{implimentation_details}
 The experiments are conducted on machines with 8 NVIDIA GTX 1080 TI GPUs.  We adopt the widely used public image datasets and models in the literature, including  CNN model for CIFAR-10 \cite{krizhevsky2009learning} and VGG-16 \cite{simonyan2015very} model for ImageNet \cite{deng2009imagenet} datasets, respectively. Results on CIFAR-10 dataset are summarized in Table \ref{table_cifar_general} and detailed in Appendix\ref{appendix:result_cifar}.
 
Unless specified, the same set of hyper-parameters is used for generating C-examples on the same dataset in our experiments. To control the trade-off between robustness and transferability, we set the weight perturbation bound $\delta$ to 0.001, 0.003, 0.005, 0.007 separately for ImageNet dataset and 0.01, 0.03, 0.05, 0.07 for CIFAR-10 dataset. 
For each C-examples generation method, 100 C-examples are generated (with randomly picked target labels) with a total of 500 iteration steps (i.e., $t=0,1, ..., 499$ as in Eq. (\ref{eqn:pgd})). 
% \PY{(make ref to eq. (3))}.
We visualize the generated C-examples on ImageNet dataset in Figure \ref{fig:visualize_img1}.

In our experiment, we use the accuracy of the C-examples on the pruned model to indicate its robustness and the accuracy on the  variant model (with  similar functionality to the base model, e.g. VGG-19 model to the base VGG-16 model) to indicate its transferabilty. Originally, the accuracy of all kinds of C-examples on the base model is  100\% during generation.
To effectively evaluate the trade-off between robustness of the pruned models and transferability to other variant models,  we define the difference between the robustness and transferability %to the worst-case model 
as \textit{Uniqueness Score} ($\textit{Uniqueness Score } = \textit{Robustness} - \textit{Transferability} $),
% \PY{(Somehow I feel uniqueness = robustness - tranferability is easier to be understood)}
where higher \textit{Uniqueness Score} means the C-examples are more robust to pruned models and less transferable to variant models.
%achieve better trade-off between robustness and transferability, and vice versa. 
Intuitively, a better fingerprint method should achieve higher uniqueness score.
\textit{Uniqueness Score} can also be used to indicate the false positive problem, i.e., if \textit{Uniqueness Score} is negative, the corresponding fingerprint method is prone to make false model claims.

\subsection{Comparative Methods}
\label{comparative_methods}
There are two works that are  most relevant to our paper. \cite{le2019adversarial} extracts adversarial examples  to watermark neural networks.  Their experiment was conducted on MNIST dataset \cite{lecun1998gradient} which only contains binary images of handwritten digits. 
Although, the method in \cite{le2019adversarial} is similar to our vanilla C-examples,  we highlight that we use random initialization instead of true data and therefore our method is data-free. 
%We apply similar principle for our vanilla C-examples as in this paper, \PY{but for fair comparison we use random initialization instead of true data.}
% \textcolor{red}{[This sentence give me a feeling that our vanilla C-example is not by us, is from that paper.]} \textcolor{orange}{(revised, highlight the difference)} 
%But instead of initializing with images from testing dataset, we use random initialization.
In our experiments, we report the performance of the vanilla C-examples as a baseline rather than  the watermarking method \cite{le2019adversarial} due to their similarity.
%the basic C-examples can be used to represent the effectiveness of using adversarial examples.
Another work proposes sensitive examples \cite{he2019sensitive} from a DNN as its fingerprints. 
Similar to \cite{le2019adversarial}, its fingerprinting  also relies on  adversarial examples. 
This paper regards all the pruned models as compression attack and reject the pruned models %from the base model 
even the test accuracy  degradation after pruning is minor (e.g., 0.65\%). 
Different from \cite{he2019sensitive}, we believe that an effective fingerprinting method should be robust  to  pruned models and recognize pruned models as non-attack. % and not suffer from the false positive problem. 
To demonstrate the robustness problem of \cite{he2019sensitive}, we use pruned models to evaluate the robustness of sensitive examples.
%evaluate sensitive examples with uniqueness score using pruned models and . 
With 8 sensitive samples, the \textit{Robustness} (i.e.,  accuracy on pruned models) is only 0.04\%, demonstrating that pruning is treated as illegitimate by sensitive samples, which is unreasonable due to the wide application of DNN pruning for size reduction and inference acceleration especially on edge devices with limited resources.

\subsection{Experiment Results on CIFAR-10 Dataset}\label{appendix:result_cifar}

We use a CNN model  (referred as CNN-1) to generate C-examples from CIFAR-10 dataset. The CNN-1 model has 13 convolutinal layers and 3 fully-connected layers and can achieve an accuracy of 80.5\% on  test set. To test the transferability, we apply 20 variant CNN models with an average accuracy of 80.4\% on the test set. They share  the same model architecture as the base CNN-1 model but are trained from different random initialized weights. We summarize the experimental results for C-examples on CIFAR-10 dataset in Table \ref{table_cifar_general}. We observe that LTRC-examples with $k=2$ and $\delta = 0.03$ achieve the best uniqueness scores than other methods.

\begin{table*}[t]\small
 \centering
%  \label{table_cifar_general}
  \caption{\textbf{Uniqueness Score of C-examples on Implemented Models by Different Weight Pruning Methods on the Base model CNN-1 with CIFAR-10 Dataset:} The base model has 80.5\% accuracy on test set. The base model is pruned by unstructured pruning \cite{han2015learning} with various pruning ratio, where it is pruned for 5 times at each pruning ratio and the average accuracy degradation for pruning ratio 80\%, 90\%, and 95\% are 0.2\%, 0.2\%, and 0.8\%, respectively. Here we use an optimal setting for LTRC-examples with $k=1, 2, 3$. The robustness at each pruning ratio is reported by the summation of \textit{Uniqueness Score} and the averaged accuracy of 20 variant CNN models representing the transferability of each group of C-examples.}
  \label{table_cifar_general}
  \scalebox{0.82}[0.82]{
  \begin{threeparttable}
    \begin{tabular}{ccccccc}
    \hline
    
    % Method & \delta & \makecell{Pre-trained Model\\\textbf{CNN-1} (\%)} & \makecell{Transferability \\to \textbf{CNN-2} (\%)} & \multicolumn{3}{c}{Uniqueness Score (\%)} & & & & \makecell{80\% Pruned (\%)} & \makecell{90\% Pruned (\%)} & \makecell{95\% Pruned (\%)} \\ \hline 
    \multirow{2}{*}{Method} & \multirow{2}{*}{$\delta$} & Base Model & Transferability to& \multicolumn{3}{c}{Uniqueness Score (\%)} \\
    &  & \textbf{CNN-1} (\%) &  \textbf{Variant CNNs} (\%) & 80\% Pruned & 90\% Pruned  & 95\% Pruned  \\ \hline 
    
     Vanilla C-Example &  0 & 100 & 68 & +17 & +5 & -30 \\ \hline 
     \multirow{4}{*}{RC-Example} &  0.01 &100  & 74 & +26 & +18 & -5  \\ 
     &  0.03 & 100 & 78 & +22 & +16 & +2  \\ 
     &  0.05 & 100 & 94 & +6 & +6 & +3 \\ 
     &  0.07 & 100 & 96 & +4  & +4 & +2  \\ \hline
    %  \multirow{4}{*}{RC-Example+GM}  
    %  &  0.001 & 100 & 55 & +34 & +37   &  +35 \\
    %  &  0.003 & 100 & 67 & +30  &  +38 & +38 \\ 
    %  &  0.005 & 100 & 85 & +15   & +15   & +15    \\ 
    %  &  0.007 & 100 & 100 & +0 & +0  & +0  \\ \hline 
    %  \multirow{4}{*}{LT-RCE(k = 5)）} &  & 0.001 & 100\% & 26\% & 79\% & 78\%   &  88\% & 71\% & 40\%\\ 
    %  &  & 0.003 & 100\% & 68\% & 94\%  &  96\% & 96\%  & 94\%  &  80\%  \\ 
    %  &  & 0.005 & 100\% &90\% & 98\%   & 98\%  & 96\%  & 95\%  & 86\%  \\ 
    %  &  & 0.007 & 100\% & 93\% & 100\% &100\%  & 100\% & 100\% & 99\%  \\ \cline{1-10}
    %  & \multirow{4}{*}{LT-RCE(k = 10)）} & 0.001 &  &  &  &  &  &  &  \\ 
    %  &  & 0.003 &  &  &  &  &  &  &  \\ 
    %  &  & 0.005 &  &  &  &  &  &  &  \\ 
    %  &  & 0.007 &  &  &  &  &  &  &  \\ \hline
    %  & \multirow{4}{*}{LT-RCE(k = 20)）} & 0.001 &  &  &  &  &  &  &  \\ 
    %  &  & 0.003 &  &  &  &  &  &  &  \\ 
    %  &  & 0.005 &  &  &  &  &  &  &  \\ 
    %  &  & 0.007 &  &  &  &  &  &  &  \\ \hline
     \multirow{5}{*}{\makecell{LTRC-Example\\(k = 1)}} 
     &  0 & 100 & 59 & +39  & +20 & -5  \\ 
     & 0.01 & 100 & 64 & +35 & +18 & -9  \\ 
     &  0.03 & 100 & 78 & +21 & +15 & +1  \\  
     &  0.05 & 100 & 80 & +11 & +6 & -9 \\  
     &  0.07 &100  & 83 & +10 & +5  & -3 \\ \hline
     \multirow{5}{*}{\makecell{LTRC-Example\\(k = 2)}}  
     &  0     & 100 & 28 & +42 & + 40 & +35  \\ 
     &  0.01 & 100 & 31 & +48 & +44  &  +38 \\
     &  0.03 & 100 & 51 & \textbf{+49}  & \textbf{+48} & \textbf{+43}  \\ 
     &  0.05 & 100 & 61 & +39  & +36  & +35  \\
     &  0.07 & 100 & 69 & +31 & +31  & +30 \\ \hline
     \multirow{5}{*}{\makecell{LTRC-Example\\(k = 3)}} 
     & 0 & 100 & 36& +20  & +19 & +15  \\ 
     & 0.01 & 100 & 39 & +25 & +21 & +17  \\ 
     &  0.03 & 100 & 60 & +15  & +13 & +9  \\  
     &  0.05 & 100 & 71 & +13 & +9 & +4 \\  
     &  0.07 &100  & 75 & +12 & +9 & -3\\ \hline
    \end{tabular}
  \begin{tablenotes}
          The experiment is evaluated on 100  examples generated from base model CNN-1.
  \end{tablenotes}
\end{threeparttable}}
\end{table*}

\end{document}